\documentclass{article} 
\usepackage{iclr2024_conference,times}
\usepackage{makecell}
\usepackage{tabularx}

\usepackage{amsmath,amsfonts,bm}

\def\eqref#1{equation~\ref{#1}}

\def\1{\bm{1}}

\DeclareMathAlphabet{\mathsfit}{\encodingdefault}{\sfdefault}{m}{sl}
\SetMathAlphabet{\mathsfit}{bold}{\encodingdefault}{\sfdefault}{bx}{n}

\usepackage{setspace}
\usepackage[hidelinks]{hyperref}
\usepackage{fontawesome}
\usepackage{xcolor}

\hypersetup{
    colorlinks,
    linkcolor={magenta},
    citecolor={cyan},
    urlcolor={magenta}
}

\usepackage{url}
\usepackage{colortbl}

\usepackage{amsmath,bm}
\usepackage{graphicx}
\usepackage{microtype}
\usepackage{amsfonts}
\usepackage[noend]{algpseudocode}
\usepackage{algorithmicx,algorithm}
\usepackage{color}
\usepackage{etoolbox,lipsum}
\usepackage{float}
\usepackage{caption}
\usepackage{booktabs}
\usepackage{multirow}
\usepackage{array}
\usepackage{xspace}
\usepackage{enumerate}
\usepackage{nth}
\usepackage{cleveref}
\usepackage{bbm}
\usepackage{longtable}
\usepackage{CJKutf8}
\usepackage{tcolorbox}
\usepackage{xcolor}
\tcbuselibrary{breakable}

\title{Training Report of TeleChat3-MoE}

\begin{document}

\maketitle

\begin{abstract}
TeleChat3-MoE is the latest series of TeleChat ~\cite{telechat,wang2024telechat,wang2025telechat2,li202452b,li2024tele} large language models, featuring a Mixture-of-Experts (MoE) architecture with parameter counts ranging from 105 billion to over one trillion, trained end-to-end on Ascend NPU cluster~\cite{huawei-ascend910b}. This technical report mainly presents the underlying training infrastructure that enables reliable and efficient scaling to frontier model sizes. We detail systematic methodologies for operator-level and end-to-end numerical accuracy verification, ensuring consistency across hardware platforms and distributed parallelism strategies. Furthermore, we introduce a suite of performance optimizations, including interleaved pipeline scheduling, attention-aware data scheduling for long-sequence training, hierarchical and overlapped communication for expert parallelism, and DVM-based operator fusion. A systematic parallelization framework, leveraging analytical estimation and integer linear programming, is also proposed to optimize multi-dimensional parallelism configurations. Additionally, we present methodological approaches to cluster-level optimizations, addressing host- and device-bound bottlenecks during large-scale training tasks. These infrastructure advancements yield significant throughput improvements and near-linear scaling on clusters comprising thousands of devices, providing a robust foundation for large-scale language model development on hardware ecosystems.
\end{abstract}

\begin{table}[htbp] \small
\begin{center}
\renewcommand{\arraystretch}{1.3}
\begin{tabular}{ll}
\toprule
     \textbf{Model Series} & \textbf{Github Link}  \\ 
\midrule
 \rowcolor{pink!25}\textbf{TeleChat3} &   \url{https://github.com/Tele-AI/TeleChat3}  \\

\bottomrule
\end{tabular}
\end{center}

\vspace{-2mm}
\end{table}

\newpage
\tableofcontents

\section{Introduction}
\label{sec:intro}

The scaling of large language models (LLMs) continues to drive performance improvements in natural language processing~\cite{an2025ai,shao2025ai,kaplan2020scaling,chen-etal-2025-revisiting, ngo2025symmetry,xiong2025tablezoomer,li2025mr,wu2025multi}, with Mixture-of-Experts (MoE) architectures offering an efficient path to extreme parameter counts while limiting active computation~\cite{lepikhin2020gshard,fedus2022switch,deepseek-v3,qwen3,kimi-k2,hu2025mosaic}. Training such models at hundreds of billions to trillions of parameters, however, imposes severe demands on numerical stability, distributed efficiency, and parallelism strategy optimization across large-scale accelerator clusters~\cite{shoeybi2019megatron,rasley2020deepspeed,rajbhandari2022deepspeed,narayanan2021efficient,chowdhery2022palm}.

This technical report documents the training infrastructure developed for TeleChat3-MoE, a series of large-scale MoE models from TeleAI, trained end-to-end on Huawei Ascend NPU clusters using the MindSpore framework. The design of the models employs hardware-aware architectural features---including Multi-Latent Attention (MLA, ~\cite{deepseek-v2}), shallow-and-wide topologies, and high-sparsity MoE architecture. The infrastructural advancements enable reliable and efficient training at frontier scales.

The key contributions of this work are as follows:
\begin{itemize}
  \item Systematic accuracy verification methodologies, encompassing operator-level precision checks and end-to-end alignment across hardware platforms, frameworks, and parallelism strategies.
  \item Performance optimizations in training framework, including interleaved pipeline scheduling with 1F1B communication-computation overlap, attention-aware data scheduling for long-sequence load balancing, hierarchical and overlapped communication for expert parallelism, and DVM-based cross-class operator fusion.
  \item A systematic parallelization tool that combines analytical estimation with integer linear programming to efficiently optimize multi-dimensional parallelism configurations, reducing tuning time from days to hours while achieving comparable or superior performance.
  \item Methodological cluster-level optimizations, focusing on identifying and mitigating host- and device-bound bottlenecks through profiling-driven resource isolation and firmware enhancements, ensuring consistent performance across heterogeneous nodes.
\end{itemize}

These advancements yield high Model Flops Utilization (MFU), near-linear scaling on thousands of devices, and reproducible training outcomes. By open-sourcing the models and infrastructure, this work provides a comprehensive, high-performance training stack tailored to hardware ecosystems, facilitating further progress in large-scale language model research.

\section{Model Architecture}
\label{sec:model_architecture}

The TeleChat3-MoE series of LLMs, encompassing models ranging from 105B to trillions of parameters, adopts a unified Mixture-of-Experts (MoE) architecture. To enable efficient training and inference of frontier-scale MoE models, we introduce several architectural refinements that tightly couple with our high-performance infrastructure.

Key architectural features tailored for affinity include:
\begin{itemize}
  \item \textbf{Multi-Latent Attention (MLA).} Unlike traditional Multi-Head Attention (MHA, \cite{vaswani2017transformer}) and Grouped Query Attention (GQA, \cite{ainslie2023gqa}), MLA compresses KV vectors into a low-dimensional latent space, significantly slashing KV cache~\cite{tay2020efficient} usage as well as increases computational intensity~\cite{williams2009roofline} during inference (especially for long-context), yielding a substantially higher compute-to-memory-access ratio compared to GQA.
  \item \textbf{Shallow-and-Wide Model Depth.} TeleChat3 models employ relatively fewer layers (45--60+ depending on scale) with wider hidden dimensions and intermediate sizes. This ``shallow-and-wide'' topology enhances computational density per layer, reduces pipeline bubble overhead in large-scale distributed training, and improves overall cluster MFU. Combined with our interleaved pipeline scheduling and 1F1B overlap (Section~\ref{subsec:pipeline}), it enables higher effective utilization on massive clusters.
  \item \textbf{High-Sparsity MoE with Low Activation Ratio.} TeleChat3-MoE features an aggressively sparse MoE architecture, where active parameters constitute only a small fraction of total parameters (e.g., top-4 to top-8 routing out of hundreds of routed experts plus only one shared experts). This results in low activation memory footprint relative to model size. During inference, the architecture supports significantly higher concurrency per device. In large-scale cluster deployment, it outperforms lower-sparsity MoE designs~\cite{jiang2023mixtral}. During training, through our expert parallelism optimizations (Section~\ref{subsec:ep_comm_merge} and Section~\ref{subsec:ep_comm_overlap}), we achieve superior end-to-end training throughput.
\end{itemize}

Detailed hyperparameters for representative TeleChat3-MoE models are provided in Table~\ref{tab:telechat3_arch}.

Our interleaved pipeline scheduling, hierarchical EP communication, and EP communication overlapping (detailed in Section~\ref{sec:training_optimization}) enable near-linear scaling from hundreds to thousands of devices. The automatic parallelization tool (Section~\ref{sec:auto_parallel}) further ensures optimal multi-dimensional parallelism strategies are discovered rapidly. The scalability of our infrastructure is evidenced by successful end-to-end training practices from the 105B model (trained efficiently on modest clusters) to a trillion-parameter model on 8192 devices, providing a robust foundation for training next-generation MoE models at unprecedented scales.

\begin{table}[t]
\centering
\caption{Model architecture hyperparameters of the TeleChat3-MoE series. All models use a vocabulary size of 131,072 and support context lengths up to 32K--128K tokens via optimized positional embeddings.}
\label{tab:telechat3_arch}
\begin{tabularx}{\textwidth}{
  >{\centering\arraybackslash}X
  >{\centering\arraybackslash}X
  >{\centering\arraybackslash}X
  >{\centering\arraybackslash}X
  >{\centering\arraybackslash}X
  >{\centering\arraybackslash}X
  >{\centering\arraybackslash}X
  >{\centering\arraybackslash}X
  >{\centering\arraybackslash}X
}
\toprule
Params & Layers & Hidden Size & Heads & Routed Experts & Expert Intermediate Size & Experts per Token & Shared Experts  \\
\midrule
105B  & 45     & 2,560        & 32       & 192     & 1536       & 4         & 1         \\
438B  & 54     & 5,120        & 128      & 256     & 2048       & 8         & 1         \\
1119B & 61     & 5,120        & 128      & 384     & 3072       & 8         & 1         \\
\bottomrule
\end{tabularx}
\end{table}

\section{Accuracy Verification}
\label{Accuracy Verification}

\subsection{Operator-wise Accuracy Verification}

Ensuring numerical accuracy at the operator level is critical for the reliability and reproducibility of training. Minor discrepancies in fundamental operators can propagate, leading to divergent model behavior. Precision errors in operators typically originate from several reasons:
\begin{itemize}
\item \textbf{Non-associative Floating-point Operations:} Operations like addition are non-associative under finite precision. Parallel reductions (sum, mean) in different orders (e.g., data-parallel gradient aggregation) will inevitably produce ULP-level differences~\cite{gupta2015deep,micikevicius2018mixed,impacts2024fpna}.
\item \textbf{Approximation Algorithms:} Operators using iterative methods (e.g., svd, exp) or approximation functions may have different convergence criteria or polynomial approximations across backends~\cite{gupta2015deep}.
\item \textbf{Accumulation Precision:} The choice of using float32 for accumulation versus float16/bfloat16 in mixed-precision kernels significantly impacts output precision.
\item \textbf{Non-deterministic Operations:} Certain operations, such as some fused computation or communication kernels, can introduce non-determinism, causing run-to-run variance~\cite{nvidia2022nondeterministic}.
\item \textbf{Data Type Promotion \& Casting:} Implicit or explicit casting rules between fp16, bf16, fp32 can be a source of precision loss and inconsistency.
\end{itemize}
The complexity of the root causes behind operator precision issues presents a significant challenge for their characterization and localization. To address this complexity, we propose a scientific and systematic workflow for effectively checking the operator problems.

\textbf{(1) Construct the Golden Baseline}

The golden baseline (ground truth) is constructed using a CPU reference implementation. The construction method is as follows:
\begin{itemize}
\item If the test case input is of type fp16 or bf16, the input values must be cast to fp32 for computation, and the output will also be of type fp32.
\item If the test case input is of type fp32, computation is performed directly in fp32, yielding an fp32 output. Alternatively, computation may be performed in fp64 to produce an fp64 output for higher precision.
\item If the test case input is of type hf32, the input values must be cast to fp32 for computation, resulting in an fp32 output.
\end{itemize}

\textbf{(2) Define Accuracy Tolerance Based on Accumulation Count}

When establish the accuracy tolerance standard, the number of accumulation operations within the operator must be considered. Certain operators, such as ReduceSum and AvgPool, involve underlying accumulation processes. When processing inputs with large shapes, these repeated accumulations can lead to compounded numerical errors. Therefore, when defining the precision error threshold, it is necessary for the operator developer to estimate the number of accumulation operations performed in the given test scenario. 
Different tolerance values are then set based on this count. The operator is deemed to meet the accuracy requirement if the actual error from testing falls within the corresponding specified tolerance range. Based on our practical experience, reference tolerance values for different accumulation counts are provided in Table \ref{accuracy-tolerance-for-acc-cnt}.

\textbf{(3) Check Inputs for Extreme Values}

During operator accuracy verification, it is essential to examine the operator inputs for extreme values that may impact computation. The presence of such extreme values in the inputs can cause the operator to fail accuracy requirements.
A representative scenario involves division operations within the operator's computation. When the divisor is extremely small, computational errors can be significantly amplified. For example, we encountered an issue with the prod operator (which replaces elements in a specified dimension with the product of all elements in that dimension, thereby removing the dimension). A forward pass input generated an exceedingly small value. Since the backward computation of the prod operator involves division, dividing by this tiny value dramatically magnified pre-existing—but otherwise acceptable—forward computation errors, causing the result to exceed the precision tolerance.
Therefore, to mitigate the potential impact of extreme inputs on operator accuracy, we apply a clipping operation to constrain the lower bound of each operator's input value during testing.

\begin{table}[t]
\centering
\caption{The suggested accuracy tolerance standard under different accumulation count.}
\label{accuracy-tolerance-for-acc-cnt}
\begin{tabularx}{\textwidth}{
  >{\centering\arraybackslash}X
  >{\centering\arraybackslash}X
  >{\centering\arraybackslash}X
  >{\centering\arraybackslash}X
}
\toprule
\multicolumn{2}{c}{float32} & \multicolumn{2}{c}{float16} \\
\midrule
Accumulation Count & Accuracy Tolerance & Accumulation Count & Accuracy Tolerance \\
\midrule
\textless2000     & 0.0001 & \textless2000     & 0.004 \\
2000      & 0.0002 & 2000      & 0.07  \\
10000     & 0.0004 & 10000     & 0.1   \\
20000     & 0.0008 & 20000     & 0.1   \\
200000    & 0.008  & 200000    & 0.2   \\
\bottomrule
\end{tabularx}
\end{table}

\subsection{End-to-End Model Accuracy Alignment}

During the development of the TeleChat series, precision consistency verification is a critical preparatory step before large-scale training. Model migration across heterogeneous hardware platforms, changes in distributed parallelism strategies, or shifts between inference and training settings can introduce significant numerical discrepancies. Given the high cost and complexity of running large-scale cluster experiments, ensuring accuracy alignment in advance is essential to guarantee training effectiveness and reproducibility.

Two primary scenarios necessitate precision alignment: (1) cross-hardware and cross-framework migration, where differences in computation precision, operator implementations, or framework behavior can lead to divergence; and (2) changes in distributed training strategies (e.g., combinations of parallelism dimensions and degree), where partitioning and communication may introduce subtle numerical variations.

\subsubsection{Cross-Hardware Precision Alignment}

When migrating training workflows to NPUs, various precision issues may arise, such as loss divergence, poor convergence, numerical overflow, or noticeable deviations in loss curves. These discrepancies often stem from differences in operator semantics, accumulation precision in mixed-precision kernels, optimizer implementations, or framework-specific handling of randomness.

\begin{figure}[t]
    \centering
    \includegraphics[width=\textwidth]{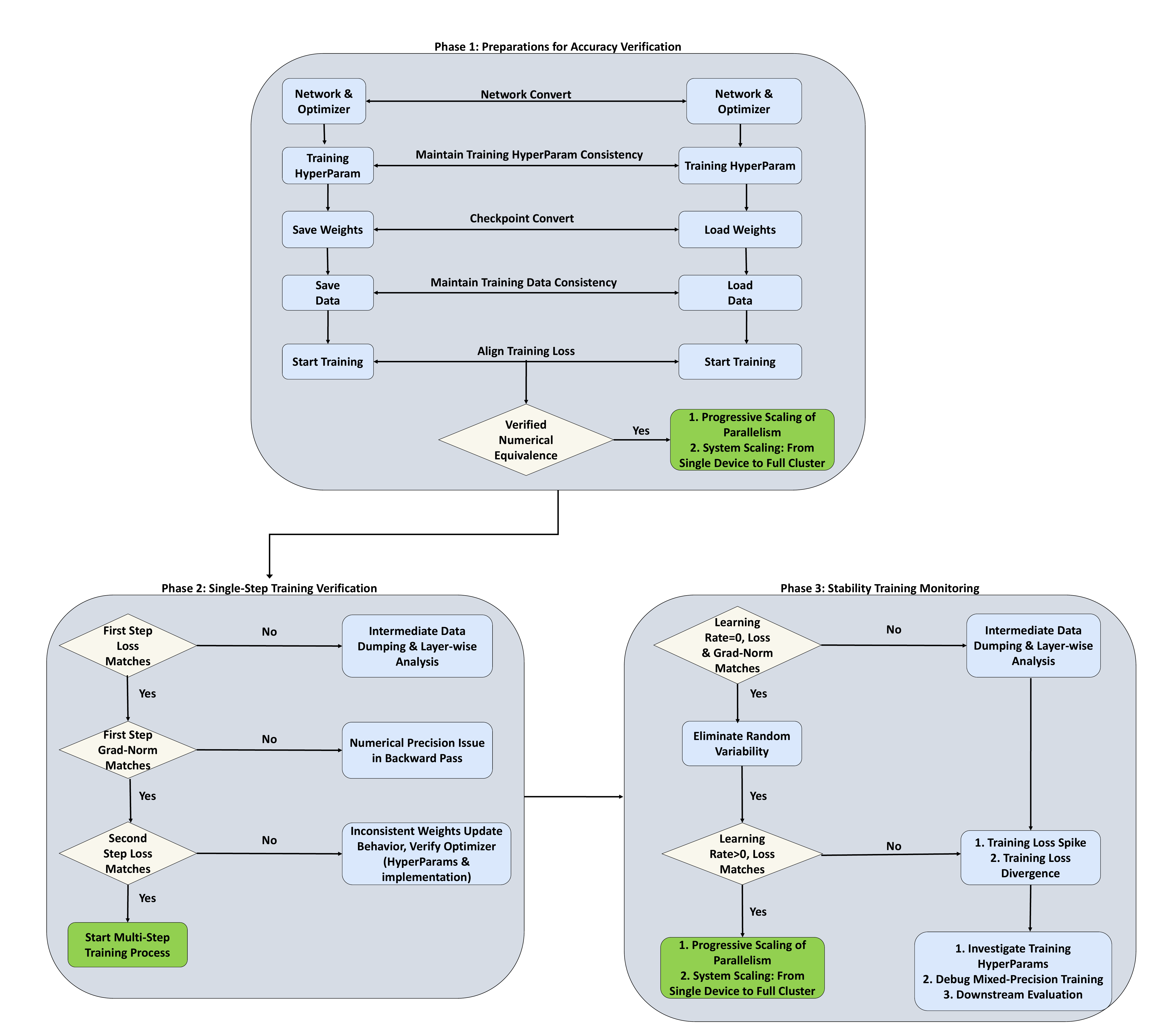}
    \caption{Workflow for cross-hardware training precision alignment.}
    \label{fig:accuracy}
\end{figure}

To systematically resolve such issues, we first ensure consistency in foundational configurations: vocabulary table (including token IDs and special symbols), network architecture (e.g., number of layers, attention heads and expert count), training hyperparameters, weight initialization/loading, mixed-precision settings, dataset preprocessing, and deterministic behavior. After these alignments, if discrepancies persist, we follow a structured localization workflow to isolate the root cause, as is shown in Figure~\ref{fig:accuracy}.

The workflow proceeds as follows:
\begin{enumerate}
    \item Start with a simplified training scenario (single-device, tiny-scale model, no parallelism) and verify initial loss and forward/backward pass equivalence.
    \item If alignment holds, incrementally scale the scenario by enabling parallelism, adding devices, and increasing model size until the discrepancy reappears.
    \item Once the problematic configuration is identified, use tensor dumping or logging tools to compare intermediate activations, gradients, or optimizer states across the heterogeneous platforms.
\end{enumerate}

This systematic approach efficiently narrows down the source of precision errors. For example, a significant deviation in initial loss typically indicates issues in forward computation (e.g., operator precision or model structure misalignment), whereas discrepancies emerging after the first backward pass often point to gradient computation or optimizer behavior.

\textbf{Practical Case:} In a large-scale MoE training project involving thousands of devices, substantial loss divergence was observed between different hardware platforms prior to the official training launch. After aligning basic model architecture, optimizer settings, weights, mixed-precision hyperparameters, and loading identical data, we performed forward pass alignment. The initial loss matched, confirming correctness of the forward computation. However, backward propagation failed to align. Tensor dumping and log comparison revealed a framework-specific inconsistency in weight decay application. Fixing this problem successfully ensured consistent loss curves across heterogeneous hardware, enabling reliable progression to full-scale training.

\subsubsection{Numerical Equivalence Validation Across Parallelism Strategies}

In the context of developing large-scale models like the TeleChat-MoE series, numerical equivalence issues may arise across varying parallelism strategies, where changes in parallelism—such as shifting from data parallelism to hybrid approaches combining tensor, pipeline, and expert parallelism—can subtly alter numerical outcomes due to differences in tensor partitioning, communication patterns, and aggregation methods.

The workflow for numerical equivalence validation mirrors the structured approach used in cross-hardware alignment but adapts it to parallelism-specific challenges. It leverages a dedicated gating cluster for systematic precision validation before and after any change in the parallelism strategy. This allows for controlled comparisons of training metrics, intermediate tensors, and optimizer states under different configurations.

The workflow proceeds as follows:

\begin{enumerate}
\item Establish baseline trajectories: Run a reference training iteration using the existing parallelism strategy (e.g., data parallelism on a small model subset) and capture key metrics such as loss values, gradients, and activation norms at predefined checkpoints.
\item Introduce the new strategy: Apply the updated configuration (e.g., incorporating tensor parallelism or pipeline interleaving) on the same dataset and hyperparameters, ensuring identical initial weights and randomness seeds for reproducibility.
\item Perform equivalence checks: Apply equivalence checking to verify consistency across forward/backward passes and optimizer steps. If the check passes, proceed to numerical validation by comparing dumped intermediate tensors and optimizer states for any deviations.
\item Iterate and localize: If discrepancies arise (e.g., mismatched gradient aggregations), incrementally revert changes to isolate the issue, such as verifying communication operators like AllReduce or ReduceScatter.
\item Scale validation: Once equivalence is confirmed at small scales, gradually increase model size, device count, and batch dimensions to ensure robustness under production loads.
\end{enumerate}

The potential impact on numerical equivalence for different parallelism strategies is drawn from our practical experience, as illustrated in Table~\ref{tab:parallelism_equivalence}.

\begin{table}[t]
\centering
\caption{Overview of parallelism strategies and potential numerical discrepancies.}
\label{tab:parallelism_equivalence}
\renewcommand{\arraystretch}{1.3}
\begin{tabularx}{\textwidth}{
  >{\raggedright\arraybackslash}p{2.8cm}
  >{\raggedright\arraybackslash}X
  >{\raggedright\arraybackslash}X
}
\toprule
\textbf{Parallelism Strategy} & \textbf{Description} & \textbf{Potential Numerical Discrepancies} \\
\midrule
Data Parallelism (DP) & Replicates model across devices; shards data batches & Gradient aggregation errors (e.g., missing AllReduce) \\
\midrule
Tensor Parallelism (TP) & Shards model weights across devices & Incorrect scaling in losses/gradients (e.g., undivided by TP degree) \\
\midrule
Pipeline Parallelism (PP) & Divides model into stages across devices; uses microbatches & Complex scheduling leading to accumulation mismatches \\
\midrule
Sequence Parallelism (SP) & Shards sequences across devices & Offset errors in positional encodings (e.g., RoPE) \\
\midrule
Expert Parallelism (EP) & Routes experts in MoE across devices & Routing/combining mismatches affecting norms \\
\bottomrule
\end{tabularx}
\end{table}


\section{Training Framework Performance Optimization}
\label{sec:training_optimization}

To achieve efficient training of large-scale models on massive clusters, we make several key optimizations in the MindSpore training framework. These optimizations target critical performance bottlenecks in pipeline parallelism, long-sequence training, Mixture-of-Experts (MoE) models, and operator execution efficiency. In this section, we describe four major improvements: interleaved pipeline scheduling with communication-computation overlap, attention-aware data scheduling for long sequences, hierarchical communication merging for Expert Parallelism (EP), and DVM-based operator fusion.

\begin{figure}[htbp]
    \centering
    \includegraphics[width=\linewidth]{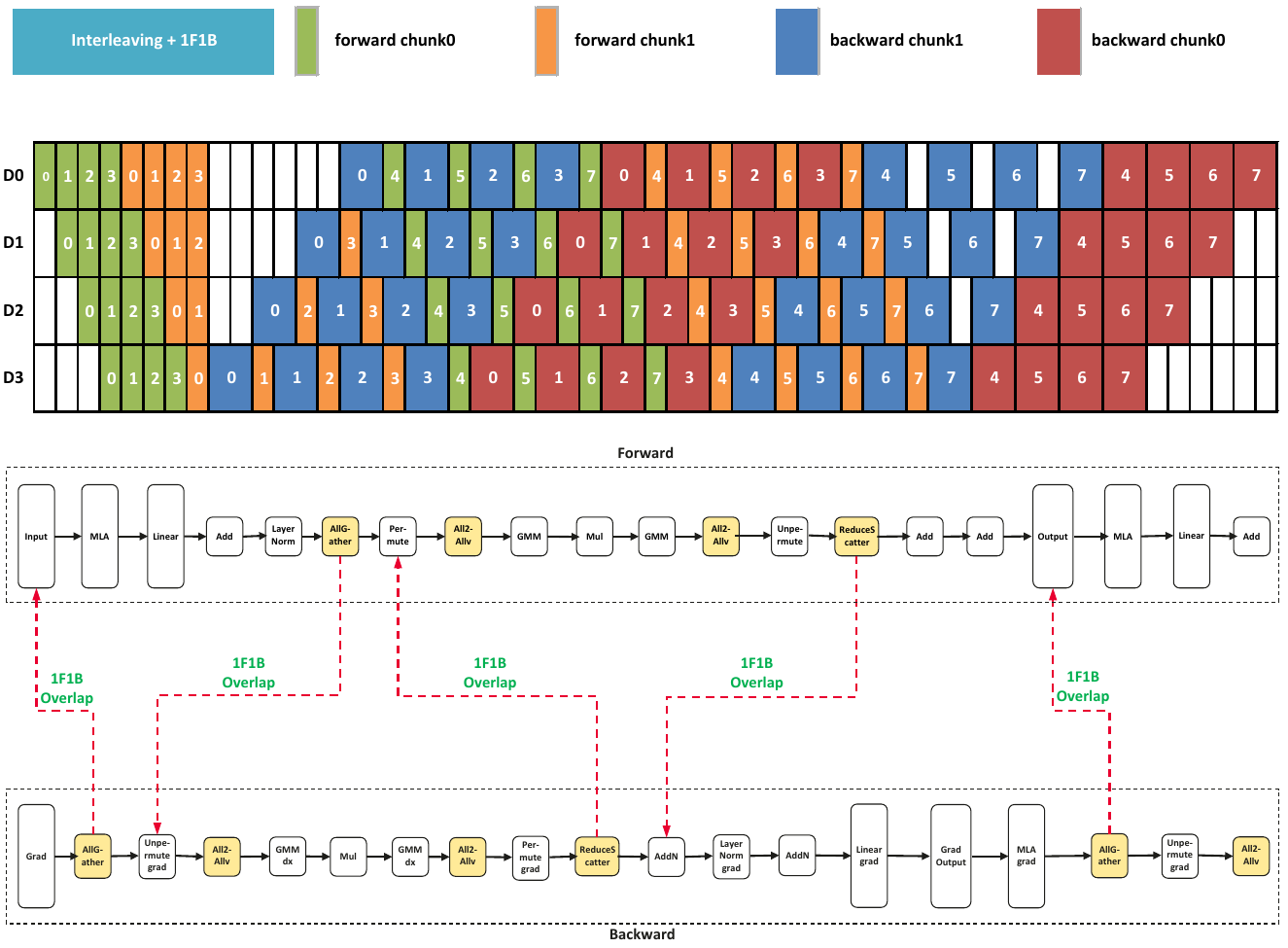}
    \caption{Interleaved pipeline scheduling with 1F1B overlap in MindSpore.}
    \label{fig:pipeline}
\end{figure}

\subsection{Interleaved Pipeline Scheduling with Communication overlapping}
\label{subsec:pipeline}

Pipeline parallelism has become a widely adopted technique for training extremely large models on large-scale clusters due to its significant memory savings and minimal communication overhead. However, a fundamental limitation of pipeline parallelism is the data dependency between consecutive stages, which introduces idle time known as ``pipeline bubbles.'' These bubbles significantly degrade training throughput, and existing pipeline schedulers, such as GPipe~\cite{huang2019gpipe} and DAPPLE~\cite{yang2020dapple}, cannot satisfy the low-bubble requirements of modern ultra-large models.

To mitigate this issue, we adopt \textbf{interleaved pipeline scheduling}, as illustrated in Figure~\ref{fig:pipeline}. Unlike conventional pipeline parallelism that assigns contiguous model layers to each stage, interleaved scheduling distributes non-contiguous layers across pipeline stages in an interleaved fashion. Although this approach slightly increases communication volume, it substantially reduces the relative communication overhead by enabling finer-grained parallelism. Building upon the original interleaved design in Megatron-LM~\cite{narayanan2021efficient},  additional memory optimizations by deferring the execution of certain forward passes. Furthermore, in the steady-state phase, we employ a carefully designed \textbf{1F1B (one forward, one backward)} overlapping strategy that schedules adjacent forward and backward passes from different micro-batches to overlap computation with communication, as is shown in Figure~\ref{fig:pipeline}.

Empirical results show that the combination of interleaved pipeline scheduling and 1F1B overlap yields an end-to-end training performance improvement of approximately 10\% compared to baseline pipeline strategies.

\subsection{Long-Sequence Optimization via Attention-Aware Scheduling}
\label{subsec:longseq}

\begin{figure}[htbp]
    \centering
    \includegraphics[width=0.5\linewidth]{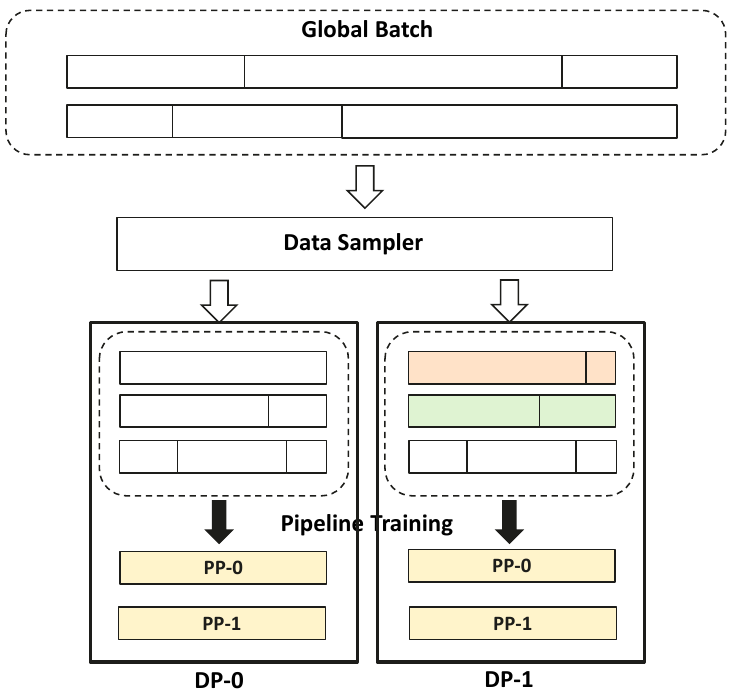}
    \caption{Attention-aware data scheduling for load balancing in long-sequence sparse attention training.}
    \label{fig:longseq}
\end{figure}

In pre-training tasks involving sparse attention with long sequences (e.g., 128K tokens), a substantial portion of devices remain underutilized due to severe load imbalance. The synchronous nature of coforces all devices to wait for the slowest device, resulting in significant throughput degradation.

The root cause is the input-dependent computational cost of sparse attention and the variability in document lengths within each training sample. Long sequences are typically concatenated from multiple shorter documents, and an End-of-Document (EoD) attention mask is applied to prevent cross-document attention, preserving long-context modeling quality. This overlapping introduces heterogeneity in per-token arithmetic intensity: tokens near the end of longer documents require substantially more attention computation. Consequently, sequences containing longer documents incur higher overall workloads despite having identical total lengths.

To address this imbalance, we propose an \textbf{attention-aware data scheduling mechanism}, depicted in Figure~\ref{fig:longseq}. By redistributing samples within each micro-batch according to the subsequence (document) lengths, the scheduler balances the total attention computation across devices. This approach achieves load-balanced sparse attention execution, substantially improves training throughput for long-sequence models, and reduces overall computational cost.

\subsection{Hierarchical Communication Merging for Expert Parallelism}
\label{subsec:ep_comm_merge}

In Mixture-of-Experts (MoE) training with Expert Parallelism (EP)~\cite{lepikhin2020gshard,rajbhandari2022deepspeed}, inter-device communication dominates training time. The intuitive implementation relied on a global All-to-All collective across all EP devices, which treats intra- and inter-device bandwidth uniformly and introduces significant communication redundancy.

\begin{figure}[htbp]
    \centering
    \includegraphics[width=0.7\linewidth]{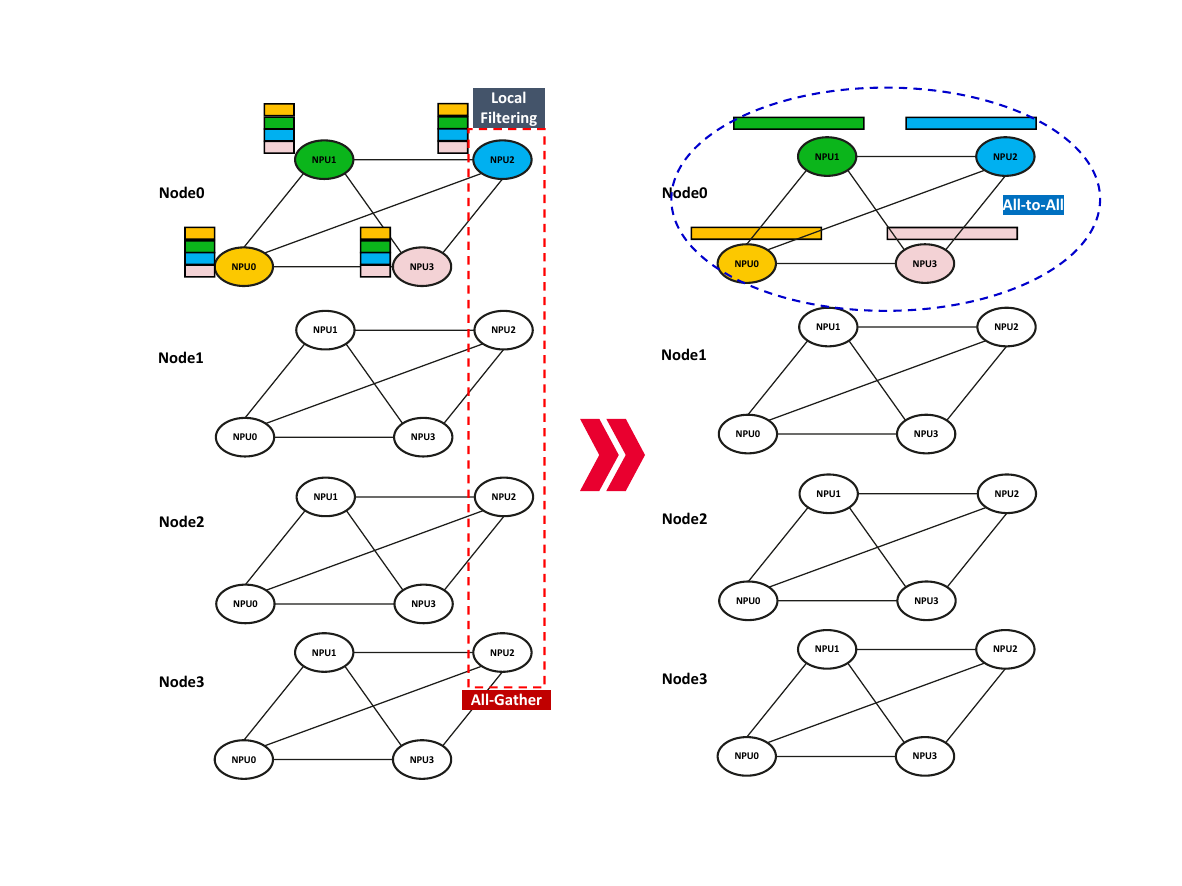}
    \caption{Hierarchical communication scheme for expert parallelism, reducing redundant cross-machine traffic.}
    \label{fig:ep_comm_merge}
\end{figure}

We introduce a \textbf{hierarchical topology-aware communication scheme}, as shown in Figure \ref{fig:ep_comm_merge}, that replaces the global All-to-All with a three-step workflow: (1) an inter-node AllGather to collect complete EP data on each machine; (2) local filtering on each node to retain only the required expert copies; and (3) an intra-node All-to-All for final redistribution among local devices. By aligning communication patterns with the physical cluster topology, this design dramatically reduces cross-device traffic.

In practice, the hierarchical scheme eliminates redundant EP communication overhead. Under an EP degree of 16, it delivers approximately 15\% higher training throughput compared to the baseline global All-to-All approach.

\subsection{Communication Overlapping for Expert Parallelism}
\label{subsec:ep_comm_overlap}

Building upon the aforementioned hierarchical Expert Parallelism communication scheme, we design an EP communication overlapping technique tailored to the network architecture of NPUs. This technique employs multi-dimensional partitioning of input data and fine-grained scheduling of computation and communication processes to overlap most of the communication overhead in the Expert Parallelism phase.

\begin{figure}[htbp]
    \centering
    \includegraphics[width=\linewidth]{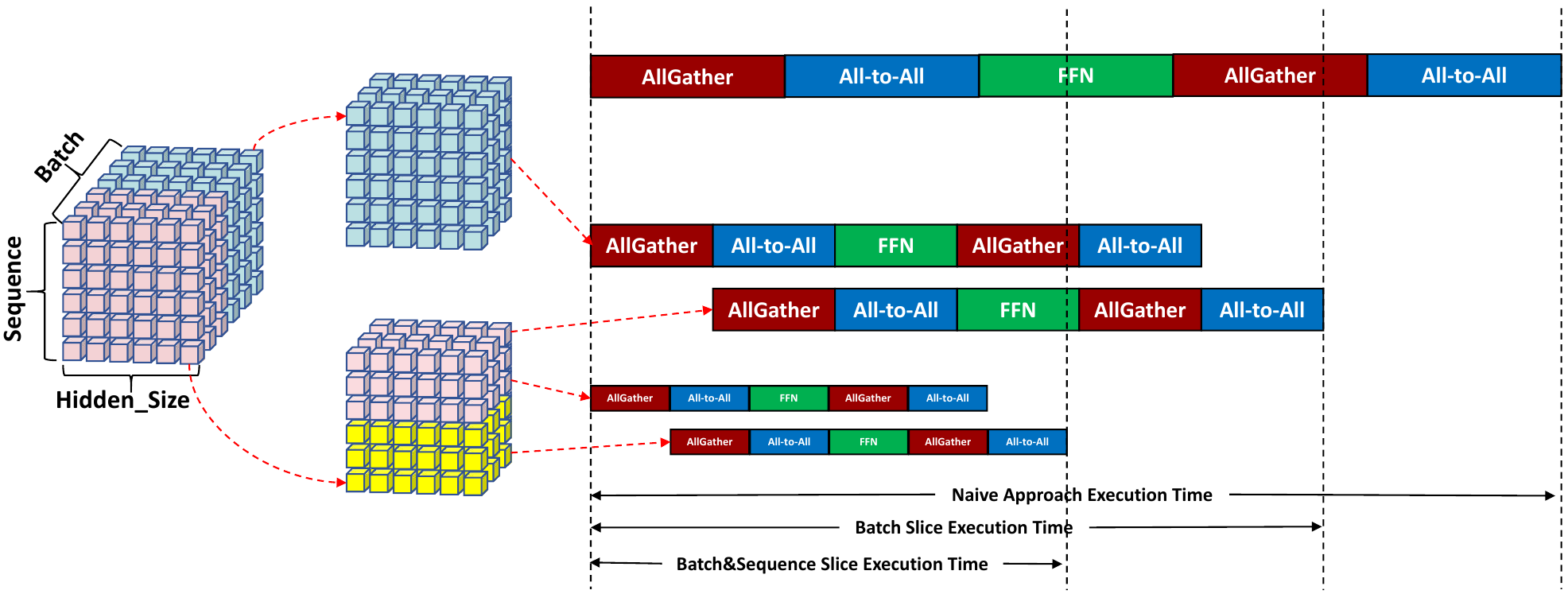}
    \caption{Expert parallelism communication overlapping via multi-dimensional data partitioning and fine-grained scheduling.}
    \label{fig:ep_comm_overlap}
\end{figure}

As illustrated in Figure \ref{fig:ep_comm_overlap}, in the context of distributed training, we maintain multiple concurrent execution data streams. The Batch and Sequence dimensions of the input data are partitioned, and the sharded data are fed into different execution streams. EP Communication in one stream can be overlapped by the FFN network computation in another stream. Furthermore, since devices perform inter-node AllGather communication over the RoCE network and intra-node All-to-All communication over the HCCS network, AllGather and All-to-All operations from different data streams ideally do not contend for bandwidth and can proceed in parallel. Therefore, through multi-dimensional data partitioning and fine-grained scheduling, our approach achieves not only computation-communication overlap but also communication-communication overlap, which significantly reduce the communication overhead in expert parallelism.

In practice, with the EP communication overlapping technique, we reduce the proportion of EP communication time in the total communication time during training from 30\% to 5\%, significantly improving the training efficiency of MoE models.

\subsection{DVM-Based Operator Fusion}
\label{subsec:fusion}

Large-scale models often contain numerous fragmented, memory-bound small operators that limit compute unit utilization due to frequent global memory accesses and kernel launch overhead.

We propose an \textbf{automatic operator fusion technique} built on the DVM (Device Virtual Machine) framework. The approach fuses memory-bound operators with neighboring operators for joint execution, eliminating explicit materialization of intermediate tensors and reducing global memory traffic. The default Vector-class fusion mechanism provides stable performance gains across a broad range of elementwise and lightweight operators.

To further address prevalent Cube-class operators (e.g., GroupedMatMul, MatMul, BatchMatMul), we extend fusion support to explicitly include Cube-class kernels and enable cross-class fusion between Cube and Vector operators. This allows concurrent execution of compute-intensive and memory-intensive kernels while leveraging on-chip L2 cache for intermediate data reuse, further alleviating memory bandwidth bottlenecks.

\begin{figure}[htbp]
    \centering
    \includegraphics[width=\linewidth]{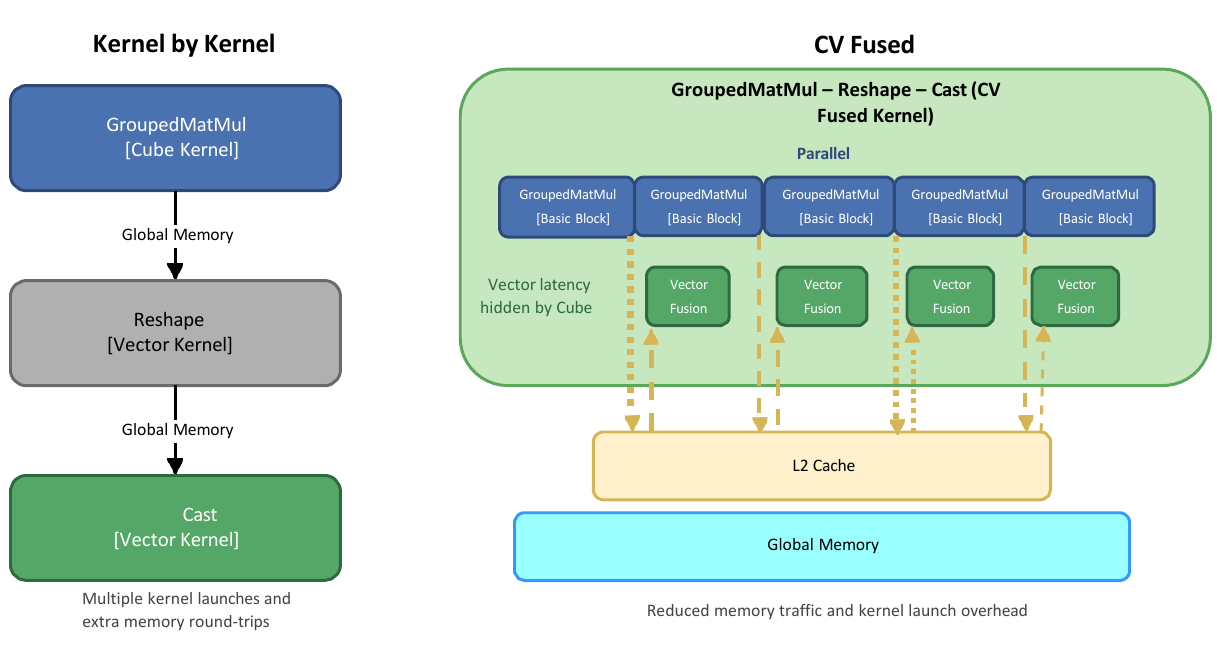}
    \caption{DVM-based cross-class operator fusion example for GroupedMatMul--Reshape--Cast sequence.}
    \label{fig:fusion}
\end{figure}

A representative example is the fusion of a common GroupedMatMul--Reshape--Cast sequence into a single composite operator, illustrated in Figure~\ref{fig:fusion}. The Cast computation is overlapped with GroupedMatMul execution, eliminating additional memory accesses. For an output tensor shape of [20, 5120, 3072], the fused kernel achieves approximately 85\% higher performance at the single-operator level compared to the unfused baseline.

These framework-level optimizations collectively enable efficient and scalable training of frontier large models on large clusters, delivering substantial end-to-end performance improvements across diverse model architectures and training configurations.

\section{Systematic Parallelization framework for Large-Scale MoE Training}
\label{sec:auto_parallel}

Training Mixture-of-Experts (MoE) models with hundreds of billions of parameters, such as our 438B-parameter model, demands careful co-optimization across multiple parallelism dimensions, including data parallelism (DP), tensor parallelism (TP)~\cite{shoeybi2019megatron}, pipeline parallelism (PP)~\cite{huang2019gpipe}, virtual pipeline interleaving (VPP)~\cite{narayanan2021efficient}, sequence parallelism (SP)~\cite{li2021sequence,liu2023ring}, expert parallelism (EP)~\cite{lepikhin2020gshard,rajbhandari2022deepspeed}, and optimizer parallelism (OP, similar to Zero)~\cite{rajbhandari2020zero}. These dimensions are tightly coupled, directly influencing per-device memory usage (parameters and activations), pipeline efficiency (bubble ratio and load balance), and communication overhead (e.g., DP gradient AllReduce, EP All-to-All/AllGather, and TP collectives). Pipeline parallelism further complicates the design space by requiring detailed stage partitioning, layer-to-stage assignment, interleaving via VPP, and selective recomputation.

In conventional expert-driven workflows, engineers rely on huristic and extensive trial-and-error to explore this vast space. The methodology is first assessed by estimating static memory against cluster size, followed by greedy pipeline configuration and recomputation decisions. Enabling VPP exponentially expands the configuration space, often leading to suboptimal choices. Each candidate typically requires memory-validation dry runs (~30 minutes) and, for promising configurations, full-scale profiling on thousands devices (~1 hour per run). Due to limited tuning time windows (often ~4 hours per session), the end-to-end manual process spans approximately 7 days and depends heavily on expert intuition.

To address these challenges, we develop a \textbf{systematic parallelization framework} that systematically explores and optimizes the multi-dimensional parallelism space. The tool (1) enumerates candidate strategies directly from the model configuration, (2) analytically estimates memory and performance to prune the search space, and (3) employs an integer linear programming (ILP) solver to jointly optimize pipeline stage assignment, VPP interleaving, and recomputation under strict per-device memory constraints. This approach reduces the tuning cycle from ~7 days to ~0.5 days while delivering throughput comparable to or better than expert-tuned baselines for 438B MoE training.

\begin{table}[t]
\centering
\caption{Comparison of expert manual tuning and the systematic parallelization framework.}
\label{tab:auto_parallel_comparison}
\begin{tabularx}{\textwidth}{lXX}  
\toprule
Aspect & Expert Manual Tuning & Strategy Generation Tool \\
\midrule
Iteration cost & Numerous dry runs + large-scale profiling & Analytical estimation + ILP; minimal dry runs \\
\midrule
Total human time & $\sim$7 days & $\sim$0.5 day \\
\bottomrule
\end{tabularx}
\end{table}

\begin{figure}[htbp]
    \centering
    \includegraphics[width=\linewidth]{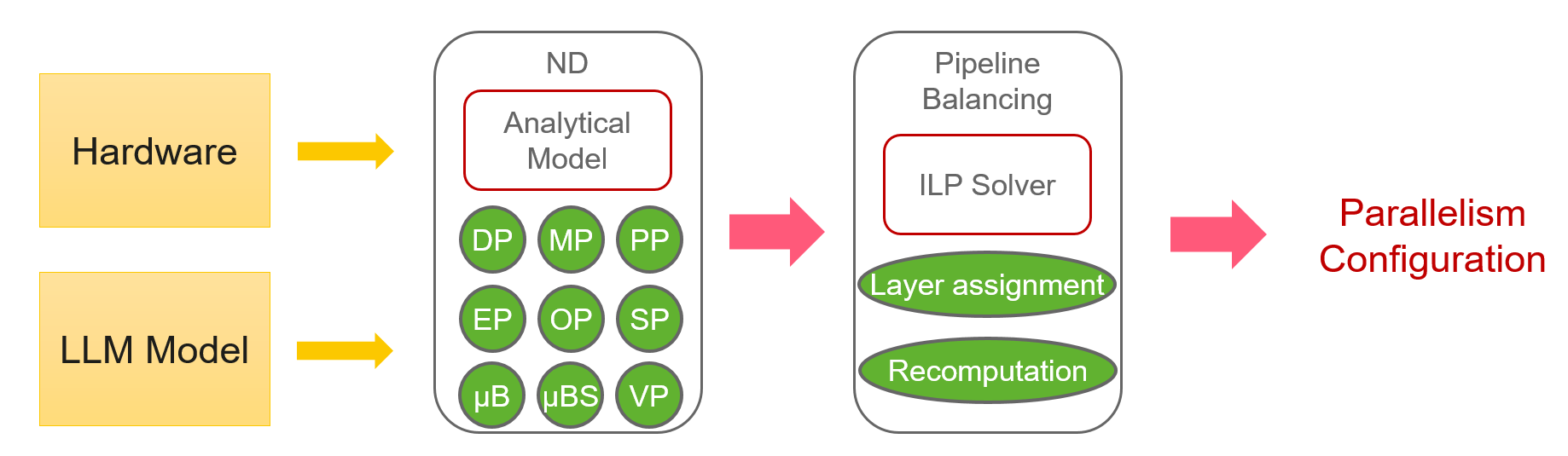}
    \caption{High-level workflow of the systematic parallelization framework.}
    \label{fig:auto_parallel_pipeline}
\end{figure}

\subsection{Conventional Expert-driven Tuning Challenges}

The conventional expert-driven workflow begins with enumerating basic parallelism strategies by sweeping combinations of DP, TP, PP, VPP, EP, OP, per-device batch size, and micro-batch count. Priority is given to PP and EP to distribute experts across machines while constraining TP to intra-node communication to avoid expensive cross-node collectives. Micro-batch count is then adjusted to minimize pipeline bubbles, followed by maximizing per-device batch size under memory limits, often trading off with recomputation.

For each candidate, engineers design pipeline stage offsets according to their expertise, layer assignments across PP and VPP stages, and recomputation scopes (e.g., flash-attention or matrix multiplication operators). Memory usage should be validated through dry runs, and promising configurations undergo large-cluster profiling to identify bottlenecks such as pipeline bubbles and communication overhead.

This iterative, non-systematic process is time-consuming (estimated ~7 days per model / cluster-configuration), cluster-intensive and heavily dependent on expert intuition, highlighting the need for systematic exploration.

\subsection{Parallelization Strategy Generation}

Our framework replaces heuristic search with a structured pipeline that achieves systematic coverage of the design space while minimizing empirical validation.

\begin{figure}[htbp]
    \centering
    \includegraphics[width=\linewidth]{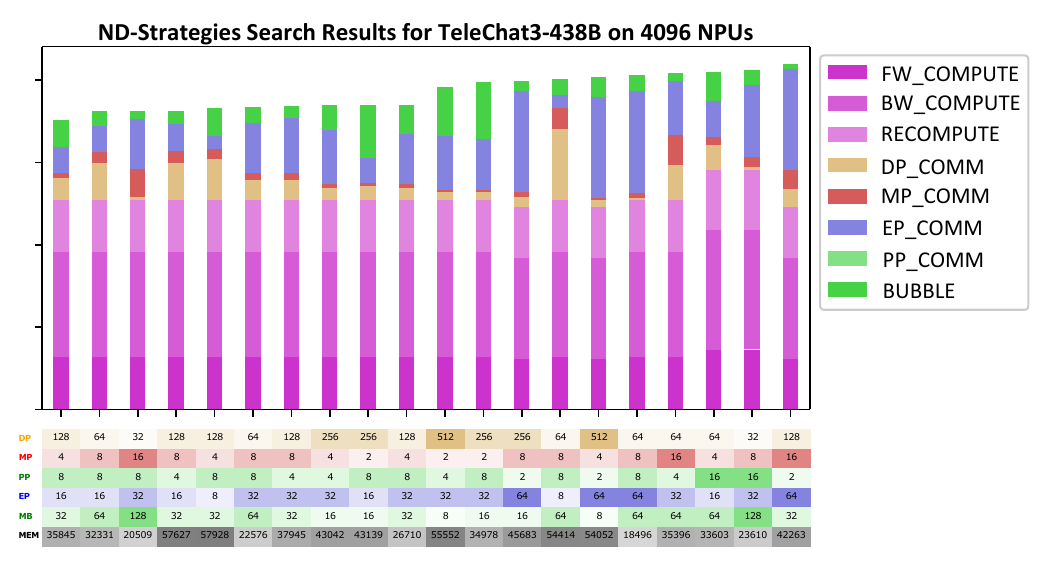}
    \caption{Example of parallelization strategy search space and estimated memory\&time-cost}
    \label{fig:auto_parallel_search_space}
\end{figure}

\textbf{High-Level Pipeline}

The workflow, illustrated in Figure \ref{fig:auto_parallel_pipeline}, proceeds as follows:
\begin{enumerate}
\item Parse the model YAML configuration to extract layer and operator details.
\item Generate candidate multi-dimensional strategies and rank them using analytical memory and performance estimates.
\item For top-ranked strategies, apply an ILP solver to optimize pipeline load balancing, including stage/chunk assignment and recomputation decisions.
\item Perform a limited number of dry runs to confirm memory feasibility.
\item Conduct final verification on the selected strategy.
\end{enumerate}

\textbf{Strategy Generation}

The tool generates ND strategies from the model YAML using configurable search limits and device counts. Taking 438B as an example, the tool performs multiple searches over the dimensions of DP, MP (tensor parallelism), PP, EP, and MB (micro-batch number) on a cluster with 4,096 devices and a global batch size of 16,384. The search results are shown in Figure \ref{fig:auto_parallel_search_space}.

The generated search space includes expert-crafted candidates, ensuring coverage of high-quality regions of the design space while enabling broader exploration.

\textbf{ILP-Based Pipeline Load Balancing}

\begin{figure}[htbp]
    \centering
    \includegraphics[width=\linewidth]{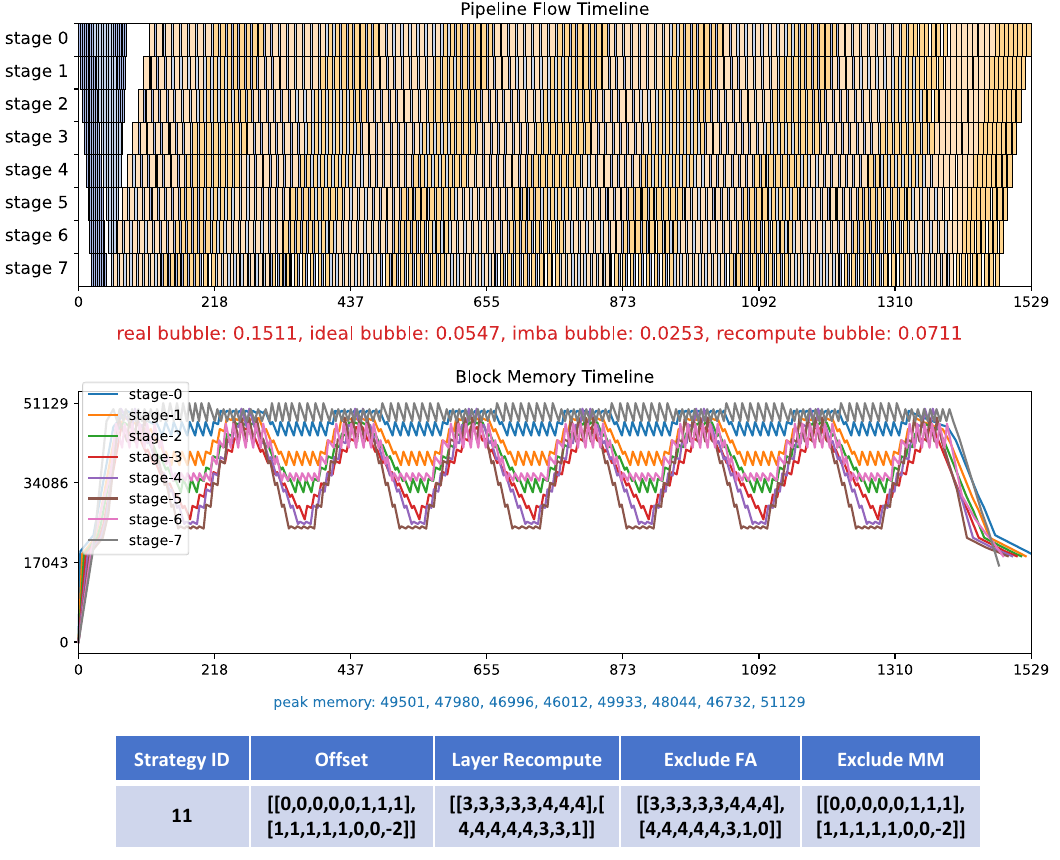}
    \caption{Example ILP-optimized pipeline stage offsets, recomputation scopes, and exclusions.}
    \label{fig:auto_parallel_ilp}
\end{figure}

For each fixed ND strategy, we optimize pipeline load balancing using symbolic formulas that estimate pipeline idle overhead (bubble) and recomputation cost. An ILP solver searches over stage/chunk assignments and recomputation configurations to jointly reduce idle time and recomputation overhead under a memory limit, yielding schedules that are typically more balanced than manual heuristics. For example, the pipeline load balancing tool performs multiple searches over the 438B strategy (which is generated through the aforementioned ND strategies searching tool) with 8 pipeline stages, 2 interleaved pipelines, 64 micro-batches and 45GB of max-device-memory-limitations. The results produced are shown in Figure \ref{fig:auto_parallel_ilp}.

\textbf{Validation Loop}

Instead of repeated large-cluster profiling for many candidates, the tool performs a small number of dry runs per shortlisted strategy to validate memory constraints. Once feasibility is confirmed, final performance verification proceeds directly.

\subsection{Evaluation}

As illustrated in Table \ref{tab:step_time_comparison}, We compare end-to-end step time on 4096 deivces. The tool-generated strategy achieves step time comparable to, or slightly better than, the best compared manual baselines under the reported settings, while substantially reducing tuning effort. By automating the exploration and optimization of the complex parallelism design space, our tool dramatically reduces engineering effort and enables faster, more reproducible scaling of frontier-scale MoE models on large clusters.

\begin{table}[t]
\centering
\caption{Step time comparison across different device scales. Lower is better. .}
\label{tab:step_time_comparison}
\begin{tabularx}{\textwidth}{
  >{\centering\arraybackslash}X
  >{\centering\arraybackslash}X
  >{\centering\arraybackslash}X
  >{\centering\arraybackslash}X
}
\toprule
Strategy & Device Num & Step time (ms)\\
\midrule

Expert-Designed-1           & 4,096   & 40,076  \\
Expert-Designed-2          & 4,096   & 40,147   \\
Tool-Generated   & 4,096   & 39,969   \\
\bottomrule
\end{tabularx}
\end{table}


\section{Training Cluster Performance Optimization}
\label{sec:cluster_optimization}

Scaling MoE models to trillions of parameters on large clusters requires addressing underlying hardware-software interactions that can introduce performance bottlenecks. We adopt a systematic methodology to identify and mitigate host-bound and device-bound issues, ensuring consistent throughput across heterogeneous nodes.

Host-bound bottlenecks arise from resource contention among processes, leading to variability in per-node performance. Profiling reveals periodic fluctuations correlated with monitoring activities and inter-process interference. To mitigate this, we employ resource isolation techniques, including CPU affinity binding and kernel-level isolation domains. These methods partition computational resources, dedicating cores to critical training processes while segregating non-essential tasks. Empirical analysis on clusters of up to 4,096 devices demonstrates that such isolation reduces variance by up to 38\% and improves mean throughput by 1-2\% in small-scale experiments, scaling to 10-15\% gains in large configurations.


Device-bound issues manifest as underutilization of devices. The Ascend NPU employs a power-saving mechanism through an "idle" mode to reduce cluster energy consumption. Specifically, a threshold is established such that when an operator's execution time persistently falls below this value, the device transitions to the "idle" state, with the clock frequency scaled down to minimize chip power dissipation. However, this approach is susceptible to erroneous triggering for certain workloads (e.g., our MoE model training, which features numerous short-duration operators in the dispatch and combine stages), resulting in unintended frequency reduction during training execution and substantial performance degradation. To mitigate the performance penalties arising from this energy-saving policy, we adjust the strategy embedded in the firmware, substantially lowering the operator execution time threshold required to exit the "idle" mode, thereby preventing frequency downscaling amid training tasks. This yields 25-30\% throughput improvements in 4096 devices tasks.

Monitoring processes, while essential for cluster health, can inadvertently degrade performance through query-induced contention. We analyze the impact of periodic queries on host-device interactions, revealing amplified latencies under strict memory management modes. Transitioning to passthrough I/O memory management unit (IOMMU) configurations reduces these overheads, eliminating significant fluctuations and yielding an additional 3-5\% gain in 4096 devices tasks.

This methodological approach---combining profiling, isolation, and firmware tuning---ensures robust, predictable performance and high utilization for deivces clusters.

\section{Conclusion}
\label{sec:conclusion}

In this report, we have detailed the training infrastructure underpinning TeleChat3, a family of large-scale MoE language models successfully scaled to trillion-parameter regimes on Ascend NPU clusters. Through systematic accuracy verification workflows, targeted optimizations---including advanced pipeline scheduling, long-sequence load balancing, hierarchical expert parallelism communication with overlap, and sophisticated operator fusion---as well as a systematic parallelization framework employing analytical modeling and integer linear programming, and cluster-level methodological optimizations addressing host- and device-bound bottlenecks, we demonstrate substantial gains in training throughput, scalability, and reproducibility.

These contributions collectively address critical bottlenecks in frontier-scale model training, establishing a mature full-stack solution tailored to domestic computational ecosystems. The resulting enhancements in hardware utilization and engineering efficiency not only enable the development of competitive TeleChat3 models but also provide a solid infrastructural foundation for future explorations of even larger and more capable language models.

The TeleChat3 models and supporting infrastructure are released openly to the research community, with the aim of accelerating progress in large language model development through reliable, efficient, and scalable training methodologies.

\bibliography{iclr2024_conference}
\bibliographystyle{iclr2024_conference}

\clearpage

\appendix

\end{document}